\pdfoutput=1

\documentclass[11pt]{article}

\usepackage{acl}

\usepackage{times}
\usepackage{latexsym}

\usepackage{pifont}
\usepackage{amssymb}
\newcommand{\cmark}{\ding{51}}
\newcommand{\xmark}{\ding{55}}
\usepackage[T1]{fontenc}

\usepackage[utf8]{inputenc}

\usepackage{microtype}

\usepackage{inconsolata}

\usepackage{tikz}
\usepackage{tabularx}
\usepackage{amsmath}
\usepackage{amsthm}
\usepackage{amssymb}
\usepackage{paralist}
\usepackage{booktabs}
\usepackage{todonotes}
\usepackage[capitalise,noabbrev]{cleveref}

\usepackage{xspace}

\newcommand*{\ie}{i.e.\@\xspace}
      
\title{Atomic Inference for NLI with Generated Facts as Atoms}

 \author{Joe Stacey\textsuperscript{1}, Pasquale Minervini\textsuperscript{2}, Haim Dubossarsky\textsuperscript{3}, \\ \textbf{Oana-Maria Camburu\textsuperscript{4}, Marek Rei\textsuperscript{1}} \\
        \textsuperscript{1}Imperial College London, \textsuperscript{2}University of Edinburgh, \\ \textsuperscript{3}Queen Mary University of London,  
        \textsuperscript{4}University College London \\
        \texttt{\{j.stacey20, marek.rei\}@imperial.ac.uk} \\
        \texttt{p.minervini@ed.ac.uk, h.dubossarsky@qmul.ac.uk, o.camburu@ucl.ac.uk} }

\date{}
\begin{document}
\maketitle
\begin{abstract}
With recent advances, neural models can achieve human-level performance on various natural language tasks. However, there are no guarantees that any explanations from these models are faithful, \ie that they reflect the inner workings of the model.
\emph{Atomic inference} overcomes this issue, providing interpretable and faithful model decisions. This approach involves making predictions for different components (or \emph{atoms}) of an instance, before using interpretable and deterministic rules to derive the overall prediction based on the individual atom-level predictions.
We investigate the effectiveness of using LLM-generated facts as atoms, decomposing Natural Language Inference premises into lists of facts.
While directly using generated facts in atomic inference systems can result in worse performance, with 1) a multi-stage fact generation process, and 2) a training regime that incorporates the facts, our fact-based method outperforms other approaches.\footnote{\url{https://github.com/joestacey/atomic_inference_anli/}} 

\end{abstract}

\section{Introduction}

Current state-of-the-art models achieve impressive performance on various natural language understanding tasks.
However, predictions from these models are not interpretable, and while existing methods can suggest plausible reasons for each prediction \cite{wiegreffe2021teach}, there are no guarantees that these reasons are faithful to the underlying decision-making process of the model \cite{lyu2023faithful, atanasova2023faithfulness}. Despite the importance of inherently interpretable models for high-stakes decision-making \cite{rudin2019stop}, few works on interpretability consider this type of model \citep{calderon2024behalfstakeholderstrendsnlp}. 

Motivated by this idea, we aim to introduce an inherently interpretable model that produces plausible and faithful explanations. This method involves decomposing an input into components (\textit{atoms}) and making hard classification decisions independently for each atom. A sequence of interpretable and deterministic rules is then applied to derive the overall prediction based on the model decisions for these atoms.
We refer to this approach as \emph{atomic inference}, producing interpretable models that reveal the specific atom-level decisions responsible for each instance-level prediction.

Atomic inference methods are effective when underpinned by an appropriate choice of atoms, allowing models to independently make accurate predictions for each component part of an input. 
We investigate the effectiveness of using generated facts as our atoms. Specifically, we use an LLM to generate a comprehensive list of facts that summarises an input. This fact decomposition results in more atoms than a sentence segmentation, providing more fine-grained model interpretability. Moreover, we show that fact-based methods can considerably outperform existing methods that use either sentences or word spans as atoms.

We test our atomic inference methods on Natural Language Inference (NLI), a task that involves reasoning about the relationship between a premise and a hypothesis. This follows previous work with atomic methods, which often consider NLI \cite{schuster-etal-2022-stretching, Joe_Logic, chen-etal-2023-propsegment} or tasks analogous to NLI \cite{glover-etal-2022-revisiting, laban-etal-2022-summac, kamoi2023wice, zhang-bansal-2021-finding, DBLP:journals/corr/abs-2312-11785, andreas_qa_natver}.
To improve our fact-based models, we introduce different strategies to make the generated fact lists comprehensive, preventing important information from being missed during inference. We further experiment with an attention-based architecture that introduces the fact-generated atoms during training. 

We describe our best performing system as FGLR (Fact-Generated Logical Reasoning), a method that achieves state-of-the-art results for atomic inference, while also outperforming several large-scale LLMs.

\section{Related Work}
Atomic inference involves making discrete, atom-level predictions that are used to determine instance-level predictions \cite{Joe_Logic, DBLP:journals/corr/abs-2312-11785}, highlighting the specific atoms that are responsible for each model prediction\footnote{For atomic inference, the possible labels for each atom do not need to align with the final task labels. For example, natural logic operators could be used as intermediate atom-level classes, similar to \citet{andreas_qa_natver}.}. This contrasts with atom-based methods that require soft atom-level predictions \citep{laban-etal-2022-summac, kamoi2023wice}, or methods where the predictions for each atom also have access to other parts of the input \citep{wu2023weakly, chen-etal-2023-propsegment, FengRecent}.

Common choices of atoms include sentences \cite{schuster-etal-2022-stretching, laban-etal-2022-summac, glover-etal-2022-revisiting}, word spans \cite{Joe_Logic, andreas_qa_natver, braun2024hypothesisdriven, krishna-etal-2022-proofver}, paragraphs \cite{glover-etal-2022-revisiting, laban-etal-2022-summac}, propositions \cite{chen-etal-2023-propsegment}, or semantic triples \cite{DBLP:journals/corr/abs-2312-11785}. Recent work has further considered the decomposition of texts into lists of facts, using language models to generate fact lists that itemise the information present \citep{kamoi2023wice, min-etal-2023-factscore}. We consider the effectiveness of using generated facts for atomic inference, with models making hard entailment decisions about each fact. 

Most atom-based methods either use existing NLI models to make atom-level predictions \citep{schuster-etal-2022-stretching, glover-etal-2022-revisiting,laban-etal-2022-summac, kamoi2023wice}, or provide additional atom-level annotations to be used for model training \citep{kamoi2023wice, chen-etal-2023-propsegment}. We choose to take a different approach, integrating the fact-level decomposition into the model training process. Following \citet{Joe_Logic}, this approach teaches models to make accurate predictions for individual facts without requiring fact-level labels during training.

We provide a direct comparison of our system with the system proposed by \citet{Joe_Logic}, which segments NLI hypotheses into spans based on the presence of nouns. This span-level approach requires models to be trained with the atoms in-the-loop, enabling span-level predictions during inference. However, when using generated facts as atoms, we can compare the performance from training with the atoms in-the-loop to using a standard NLI model to make the atom-level predictions. Unlike \citet{Joe_Logic}, we also segment the NLI premise into atoms rather than the hypothesis, requiring a different framework for both training and inference. We also introduce a range of novel fact generation strategies to avoid missing information in our generated atoms, an issue that is avoided when segmenting instances into spans.
\section{Method}
\subsection{Fact Generation}

\begin{figure}
    \includegraphics[width=\columnwidth]{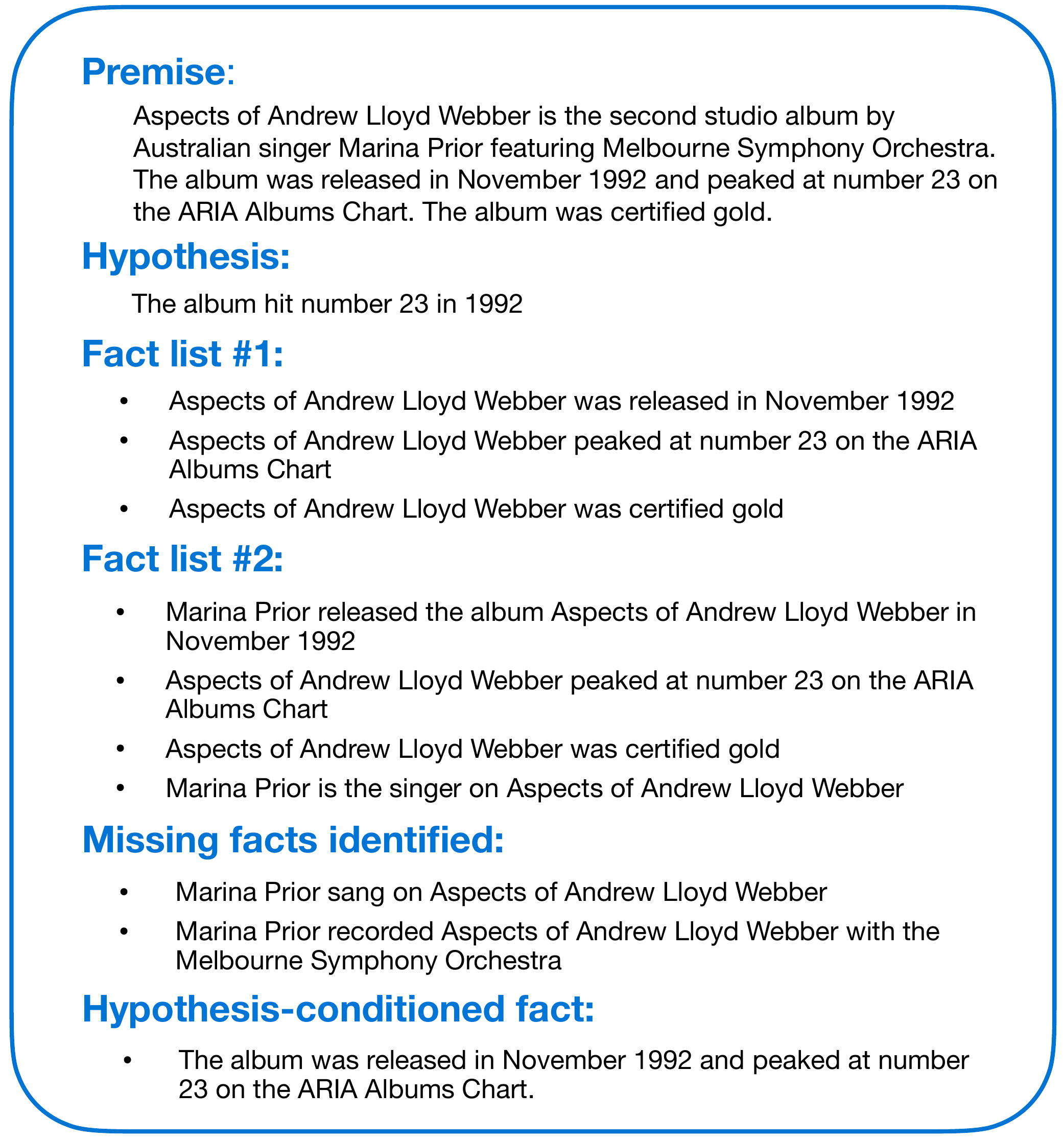} 
    \caption{Generated fact-lists for a test example, including: 1) an initially generated fact-list, 2) a second generated fact-list that can be concatenated with the first list, 3) facts that an LLM identifies are missing from the original fact list, and 4) a generated fact that is conditioned on the hypothesis.} \label{fact_gen_diagram}
\end{figure}

We define a \emph{fact} as a statement representing a single piece of information.
%
%
For each instance, we use GPT-3\footnote{See \cref{sec_appendix:modelling_details} for more information} \citep{brown2020language} to generate a fact list that itemises all of the information contained within the premise (see \cref{fact_gen_diagram}).
To generate a list of facts, we provide the language model with the premise, followed by the instruction ``List all the facts we explicitly know from the premise:''.
We implement multiple fact-generation strategies with the aim of creating more comprehensive fact lists (see \cref{fact_gen_diagram}), resulting in better performance.
This involves (1) concatenating two independent lists of facts for each NLI premise, generated using different examples in the prompt, (2) asking a generator model to extend an existing fact list, and (3) generating facts that are also conditioned on a particular hypothesis.
The hypothesis-conditioned facts are only generated for the test and validation data, so the model cannot access these facts during training.
Providing these additional facts during training would require generating considerably more facts, with a substantially higher cost.
Moreover, not providing the hypothesis-conditioned facts during training prevents models from learning from class-specific artifacts within the generated facts.

More details of the process, including an analysis of \cref{fact_gen_diagram}, are included in  \cref{sec_appendix:fact_gen_strategies_METHOD}.

\subsection{Model Architecture for Training} \label{sec:architecture}

Our FGLR (Fact-Generated Logical Reasoning) model involves an attention-based architecture that is supervised to make predictions for individual facts while only using instance-level labels for training.
This architecture has been used for token labelling \cite{DBLP:conf/naacl/ReiS18,
pislar-rei-2020-seeing, kamil} and NLI reasoning over token spans \cite{Joe_Logic}.
%
%
Each individual premise fact $i$ is encoded together with the full hypothesis using a pre-trained language model to create a representation $R_{f_i}$.
Two separate linear layers (for detecting entailment and contradiction facts) are applied to this representation to create logits for entailment ($L_{e,i}$) and contradiction ($L_{c,i}$) for each fact $i$.
To detect contradiction facts, unnormalized attention weights $\widetilde{a}_{c,i}$ are calculated as:
\begin{equation} 
\widetilde{a}_{c,i} = \sigma{(W_{c,2}(\tanh{(W_{c,1} R_{f_i} + b_{c,1})})+b_{c,2})}
\end{equation}
\noindent with parameters $W_{c,1}$, $W_{c,2}$, $b_{c,1}$ and $b_{c,2}$, with the sigmoid $\sigma$ bounding the value to a range between 0 and 1. 
These $\tilde{a}_{c,i}$ values are then normalised to create the attention distributions:
\begin{equation}
a_{c,i} = \frac{\widetilde{a}_{c,i}}{\sum_{k=1}^{m}\widetilde{a}_{c,k}}
\end{equation}
An instance-level logit $L_{c}$ is created from a weighted sum of the logits $L_{c,i}$, using the weights $a_{c,i}$.
The logit $L_{c,i}$ represents a single premise fact $i$ combined with the hypothesis, while $L_{c}$ represents a combined score for all the facts in the premise (and therefore the whole instance).
$L_{c}$ is then supervised with a loss function to predict the target label $y_{c}$ for each instance:
\begin{equation} \label{eq:obs_loss}
\mathcal{L}_{c}^{\text{Inst}} = (\sigma(W_{c,3}\times L_{c} + b_{c,3}) - y_{c})^2
\end{equation}
In addition, the unnormalised attention values $\widetilde{a}_{c,i}$ are used in a fact-level loss, encouraging the model to assign more attention to facts in contradiction examples:
\begin{equation}\label{fact_loss}
\mathcal{L}_{c}^{\text{Fact}} = (\max_{i}(\widetilde{a}_{c,i}) - y_{c})^2
\end{equation}
\noindent 
This loss indirectly teaches the model to make fact-level decisions while only using instance-level labels. The value of $y_{c}$ used in the supervision is determined by our training rules (see \cref{sec:logical_rules}).
%
%
%

The same method is then used to detect entailment facts, with parameters 
$\{ W_{e,j}, b_{e,j} \}_{j \in (1, 2, 3)}$, using the same representations $R_{f_i}$ that are used in the contradiction fact detection.
The different losses are then combined into \begin{equation}
\mathcal{L} = \mathcal{L}_{c}^{\text{Fact}} + \mathcal{L}_{e}^{\text{Fact}} + \lambda ( \mathcal{L}_{c}^{\text{Inst}} + \mathcal{L}_{e}^{\text{Inst}})
\end{equation}
using a hyper-parameter $\lambda$.

\subsection{Rules for Training and Evaluation} \label{sec:logical_rules}
The model architecture requires rules to determine the class labels during training, while also applying a second set of deterministic rules during inference.
As the FGLR method decomposes the NLI premise into atoms rather than the hypothesis, the rules introduced by \citet{Joe_Logic} are no longer applicable.
We therefore require a new set of rules compatible with the model architecture. These rules are directly compared to the rules introduced by \citet{Joe_Logic} in \cref{sec_appendix:logical_rules_comparison}.
The rules for training our method state that if an instance has a contradiction label, at least one model-generated fact must contradict the hypothesis.
Similarly, if an instance does not have a contradiction label, then none of the model-generated facts contradict the hypothesis.
The rules also state that if an example has an entailment label, then at least one of the model-generated facts must imply the hypothesis.
This involves supervising our contradiction attention layer with $y_{c}=0$ for entailment and neutral examples, and $y_{c}=1$ for contradiction examples.
For entailment examples, we supervise with $y_{e}=1$, while for neutral examples, we use $y_{e}=0$.
We do not supervise the entailment attention layer for contradiction examples\footnote{We experimented with supervising $y_{e}=0$ for contradiction examples but this marginally decreased accuracy.}.
To apply this model at an instance level during inference, we make predictions only based on the values $\widetilde{a}_{c,i}$ and $\widetilde{a}_{e,i}$ for each fact $i$.
If any $\widetilde{a}_{c,i}$ value is greater than 0.5 for any fact, then the instance is classified as a contradiction.
Otherwise, if any $\widetilde{a}_{e,i}$ value is greater than 0.5, the instance is classified as entailment.
Any instance not classified as either contradiction or entailment is predicted to be neutral.

\section{Experiments}
\subsection{Datasets} \label{sec:dataset_introduction}
We aim to introduce an atomic inference framework for challenging, multi-sentence NLI datasets where state-of-the-art models still have considerable room for improvement.
As Adversarial NLI \cite[ANLI,][]{nie-etal-2020-adversarial} exemplifies this challenge, we focus our experimentation on this dataset. 
We additionally consider out-of-distribution performance for: ConTRoL \citep{liu2020natural}, Recognizing Textual Entailment \cite[RTE,][]{wang-etal-2018-glue} and Winograd NLI \cite[WNLI,][]{DBLP:journals/corr/abs-1804-07461_GLUE, DBLP:conf/aaaiss/Levesque11_winograd}.
To avoid the baseline model needing to truncate premises, we filter ConTRoL to only include examples where the premise is < 2,000 characters. 

\subsection{Comparing Fact and Sentence Atomic Decompositions}

We compare the performance of atomic inference systems when either using generated facts, or when directly segmenting the premise into sentences. Following previous work, we test performance when making fact-level \cite{kamoi2023wice} or sentence-level \cite{schuster-etal-2022-stretching, laban-etal-2022-summac} predictions using a standard NLI model, which we train on ANLI. We update these existing methods so that they follow our atomic inference rules for evaluation, describing these approaches as FactAI (for fact atoms), and SenAI (for sentence atoms). Both FactAI and SenAI involve the same baseline NLI model trained on ANLI, with the model either making predictions for each sentence (SenAI) or for each generated fact (FactAI)\footnote{There are more facts per instance for ANLI compared to sentences, with 4.7 facts per instance on average compared to 3.0 sentences.}.

\subsection{Training with Atoms in-the-loop}

We show how fact-based methods perform better when trained with the fact atoms in-the-loop using our attention-based architecture. We additionally introduce a method of training with atoms in-the-loop using a sentence-level decomposition of the premise (which we call SenLR). This method conveniently avoids the need for a language model to generate facts, while also providing a strong comparison for our fact-based methods. Finally, we consider the performance of our fact-based model when applying alternative strategies to make the generated fact lists comprehensive. We describe our best performing system as FGLR, which involves training with a single fact list, before additionally including the hypothesis-conditioned facts during inference. 

\subsection{Baseline models}

FactAI, SenAI, SenLR and FGLR are all model-agnostic methods that can be combined with a range of uninterpretable base models. We chose DeBERTa-base \cite{he2021deberta} due to its strong performance despite having relatively few parameters (<200m). This approach exploits the strengths of both LLMs and classification models, with LLMs proving to be effective at generating fact lists, but being prone to errors in fact-level entailment decisions \citep{min-etal-2023-factscore}. We also provide further experimentation using BERT-base \cite{devlin-etal-2019-bert} and DeBERTa-large models in \cref{sec_appendix:additional_experimentation}. We directly compare our models to SENTLI \citep{schuster-etal-2022-stretching}\footnote{In the case of SENTLI we only decompose the premise, as ANLI hypotheses do not require further decomposition.} and SLR-NLI \citep{Joe_Logic}\footnote{We exclude 0.02\% of training examples due to the memory constraints of the SLR-NLI method, described in  \cref{sec_appendix:hyper_parameter_tuning}}, both atomic inference methods which we train on ANLI.
%

Finally, we compare our model performance to recent LLMs that were tested on ANLI by \citet{he2023using}, showing that models with our atom-level faithfulness guarantee can still reach or even exceed the performance of large-scale LLMs.

\begin{table*}[!t]
\begin{center}
\begin{tabular}{lcccccccc}
\toprule
 
  & \multicolumn{4}{c}{\bf In-distribution} & \multicolumn{3}{c}{\bf Out-of-distribution}\\
  \cmidrule(lr){2-5} \cmidrule(lr){6-8}
  & R1 & R2 & R3 & ANLI-all & ConTRoL & RTE & WNLI & Int? \\
\midrule
DeBERTa-base & 71.2 & 54.0 & 51.7 & 58.5 & 53.7 & 85.0 & 59.6 & \xmark \\
GPT-3.5-turbo\textsuperscript{1} & 68.5 & 54.4 & 55.9 & 59.4 & - & - & - & \xmark \\
LLaMA2 70B\textsuperscript{1} & 69.1 & 54.8 & 54.1 & 59.0 & - & - & - & \xmark \\
Mistral 7B\textsuperscript{1} & 55.5 & 43.0 & 42.5 & 46.7 & - & - & - & \xmark \\
\midrule
\textit{Span atoms:} \\
SLR-NLI\textsuperscript{2} & 65.5 & 47.8 & 47.1 & 53.0 & 48.9 & 
82.3 & 56.3 & \cmark \\
\midrule
\textit{Sentence atoms:} \\
SENTLI\textsuperscript{3} & 69.5 & 53.5 & 51.3 & 57.7 & 52.3 & 82.0 & \textbf{60.7} & \cmark \\
SenAI & 69.3 & 53.5 & 51.6 & 57.7 & 50.0 & 81.7 & 60.6 & \cmark \\
SenLR (ours) & \textbf{71.5}$\ddagger$ & \textbf{55.0}$\ddagger$ & \textbf{52.3} & \textbf{59.1}$\ddagger$ & \textbf{53.4}$\ddagger$ & \textbf{83.7}$\ddagger$ & 53.8 & \cmark \\
\midrule
\textit{Fact atoms:} \\
FactAI  & 65.2 & 50.6 & 49.9 & 54.9 & 46.6 & 77.2 & \textbf{74.6} & \cmark \\
FGLR (ours) & \textbf{71.8}$\ddagger$ & \textbf{56.1}$\ddagger$ & \textbf{55.3}$\ddagger$ & \textbf{60.7}$\ddagger$ & \textbf{49.1}$\ddagger$ & \textbf{80.8}$\ddagger$ & 70.7 & \cmark \\
\bottomrule
\end{tabular}
\end{center}
\caption{Model accuracy after training on ANLI. \textsuperscript{1} represents few-shot CoT results reported by \cite{he2023using}, while \textsuperscript{2} and \textsuperscript{3} are baselines recreated from \citet{Joe_Logic} and \citet{schuster-etal-2022-stretching} respectively using the DeBERTa base model. $\dagger$ represents results that are statistically better than the corresponding SenAI or FactAI baseline with $p < 0.05$, while $\ddagger$ represents results where $p < 0.01$, using bootstrapping statistical testing \cite{efron1993introduction}. `Int?' indicates whether the model is interpretable. All results are an average from 10 seeds.}
\label{main_results_table}
\end{table*}

\subsection{In-Distribution Results}
For basic atomic inference systems, using generated facts as atoms does not outperform a sentence atom decomposition, with SenAI outperforming FactAI for each ANLI test-set (see SenAI vs FactAI in \cref{main_results_table}). However, when training with atoms in-the-loop and including the hypothesis-conditioned facts, the FGLR system outperforms all other atomic inference methods (see \cref{main_results_table}). Training with atoms in-the-loop considerably improves performance for both sentences and fact-generated atoms, however, the benefits from this approach are greatest when using the generated facts. While interpretable models usually need to sacrifice some performance \cite{calderon2024behalfstakeholderstrendsnlp}, we find that FGLR even outperforms very large generative models. In particular, the biggest advantage of FGLR compared to other atomic inference methods is the strong performance on ANLI round-3, suggesting that fact-generated atoms help most on challenging NLI examples. 

In addition to our experimentation using DeBERTa-base, we provide results in our Appendix for implementing atomic methods with both DeBERTa-large and BERT models (see \cref{main_results_table_deberta_large} and \cref{main_results_table_bert}). We also provide a range of different ablation experiments in \cref{sec_appendix:ablation_studies} to check that all the components of FGLR are indeed necessary and beneficial. These experiments show that just under half of the improvements between FactAI and FGLR are a result of the hypothesis-conditioned facts used during inference (\cref{ablation_exp_deberta_base_3a}).

\subsection{Out-of-Distribution Results}

We identify weaknesses in atomic inference systems when testing in out-of-distribution settings, a phenomena that has not been considered in previous work. \cref{main_results_table} shows how the ANLI-trained atomic inference models perform worse than the non-interpretable DeBERTa base model for two of the three OOD datasets tested. However, we show that training with the atoms in-the-loop can help to mitigate this issue, considerably improving performance on both ConTRoL and RTE (see SenAI vs SenLR, and FactAI vs FGLR in \cref{main_results_table}).


\section{Conclusion}
We experiment with using LLM-generated facts as atoms in atomic inference systems, decomposing each NLI premise into a list of facts before making entailment predictions for each fact with the hypothesis. The instance-level predictions then depend entirely on the model's granular predictions about each fact. We find that using a standard NLI model to make predictions at a fact level results in worse performance than existing methods. However, when 1) including a multi-stage fact generation process, and 2) incorporating the generated facts during model training, our fact-based approach outperforms existing atomic inference methods. 
Our resulting FGLR model makes fact-level predictions and combines them with logical rules, without requiring fact-level labels during training. This results in high-performing, interpretable models that specify exactly which facts are responsible for each model prediction.

\section*{Limitations}

To distinguish between the entailment and neutral classes, we predict the premise as entailing the hypothesis whenever one of the individual facts from the premise implies the entire hypothesis. This approach prevents models from performing multi-hop reasoning across different premise facts. While we find that reasoning across multiple facts is unnecessary for strong performance on ANLI, there may be other datasets where this would limit performance. We propose addressing this limitation in future work.

Our method relies on decomposing the NLI premise into facts (or \emph{atoms}), determining the specific part of the input responsible for each model prediction. However, as ANLI consists of single-sentence hypotheses, no additional decomposition is required for the hypothesis. Further work would be needed to consider how to perform atomic inference over both a multi-sentence premise and a multi-sentence hypothesis when training with atoms in-the-loop.

Additionally, while our model provides interpretable decisions at an atom level, each atom-level decision itself is not interpretable. This method enables strong performance on NLI, while also providing faithfulness guarantees for the atom-level predictions.

Finally, we use GPT-3 to generate facts for each premise, which for this work cost $\sim$400 USD. As a result, we focused our experimentation on ANLI, while also including out-of-distribution evaluation on the RTE, WNLI, and ConTRoL datasets.

\section*{Acknowledgements}

We would like to thank Greg Durrett, Rami Aly and Derek Chen for all their valuable feedback on this work.

Joe Stacey was supported by the Apple Scholars in AI/ML PhD fellowship. Oana-Maria Camburu was supported by a Leverhulme Early Career Fellowship. Pasquale was partially funded by the European Union’s Horizon 2020 research and innovation programme under grant agreement no. 875160, ELIAI (The Edinburgh Laboratory for Integrated Artificial Intelligence), EPSRC (grant no. EP/W002876/1), an industry grant from Cisco, and a donation from Accenture LLP. Haim was supported by the Riksbankens Jubileumsfond (under reference number M21-0021, Change is Key! program).

\bibliography{anthology, custom}

\appendix
\clearpage

\section{Modelling Details}\label{sec_appendix:modelling_details}
We perform the fact generation for our fact lists with text-curie-001, using the same model when extending these lists of facts. However, we used \emph{text-davinci-003} when generating the facts conditioned on the hypothesis, finding that generating the facts conditioned on the hypothesis was a more difficult task. As further out-of-distribution experiments were conducted when these models were no longer available, facts were generated for the RTE and WNLI datasets using GPT-3.5-turbo \citep{brown2020language, InstructGPT}.

\begin{figure*}
    \includegraphics[width=460pt]{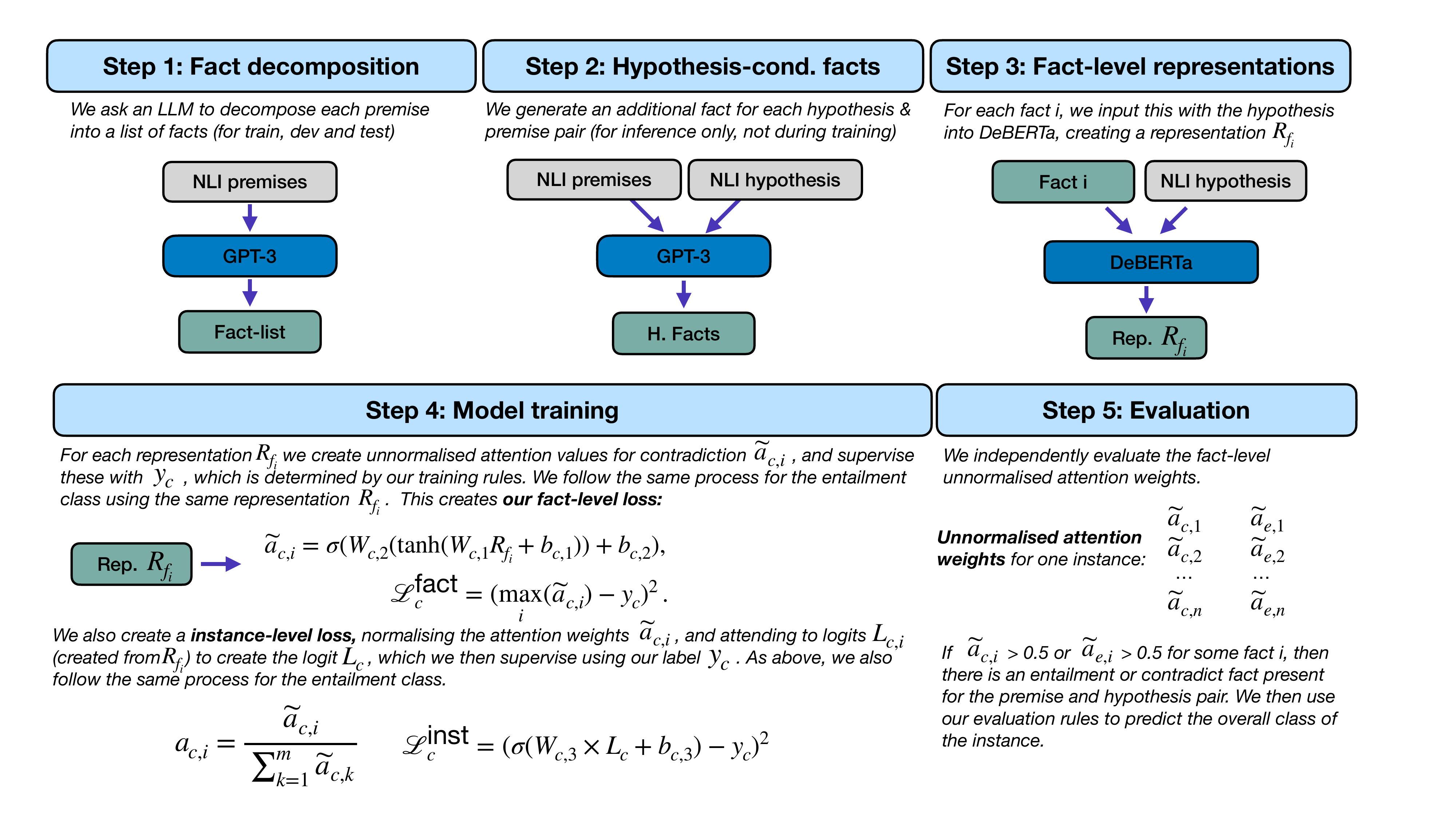} 
    \caption{Our FGLR method is summarised above, with five stages: 1) generating fact lists for each premise, 2) generating an additional fact when performing inference, prompting GPT-3 to create a relevant fact from the premise for a specific hypothesis, 3) creating a representation for each fact, 4) our fact-level and instance-level losses used in training, and 5) evaluation using our evaluation rules.} \label{FGLR_model_diagram}
\end{figure*}

For our base models, we used bert-base-uncased, deberta-v3-large, and deberta-v3-base, implemented from HuggingFace \cite{wolf2020huggingfaces}. All statistical testing was performed using a bootstrapping hypothesis test \cite{efron1993introduction}. A diagram describing FGLR can also be found in \cref{FGLR_model_diagram}.


\section{Logical Rules Comparison} \label{sec_appendix:logical_rules_comparison}
\begin{figure*}
    \includegraphics[width=460pt]{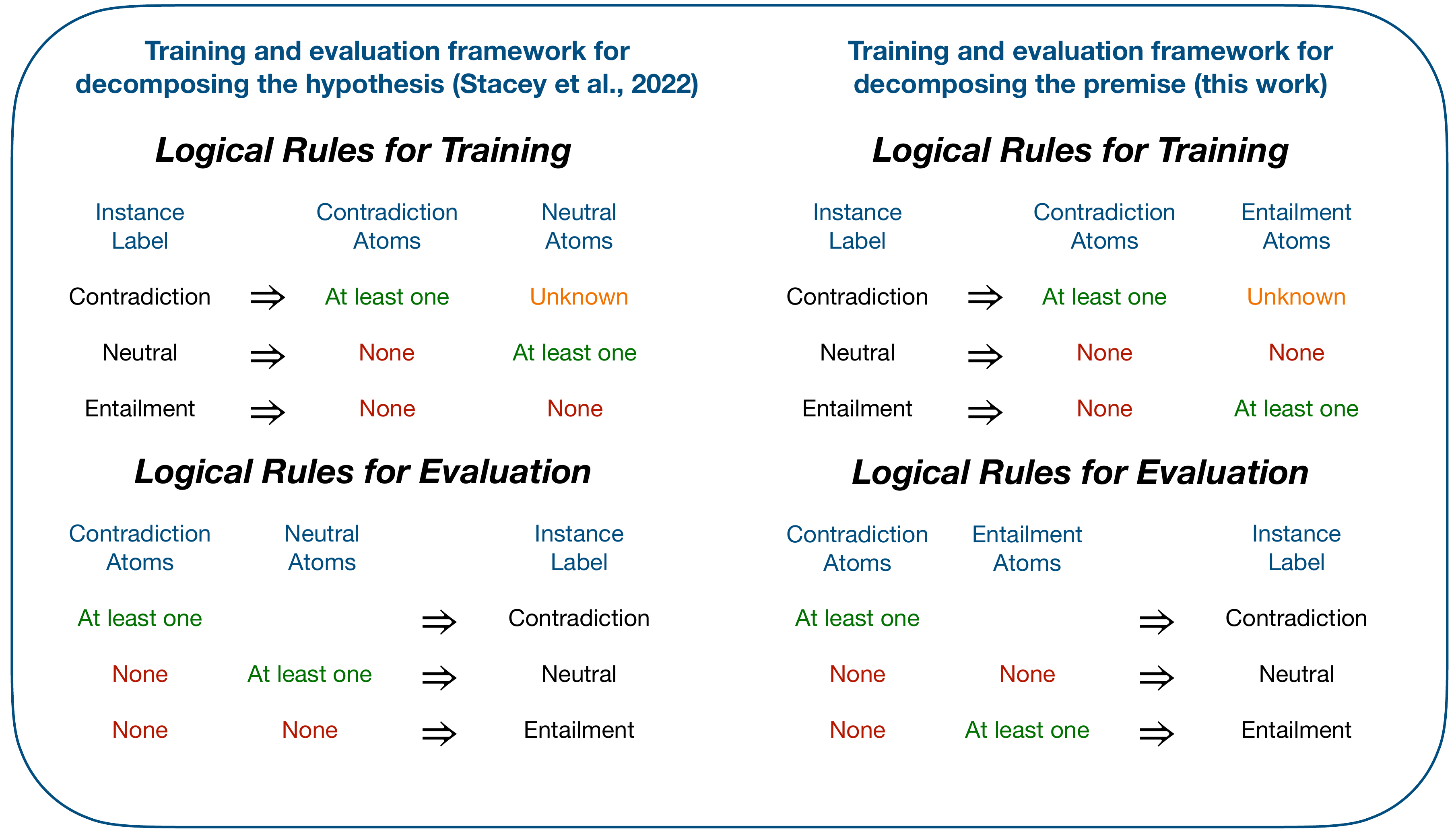} 
    \caption{Inference and training framework from our work decomposing the NLI premise, compared to the rules used by \citet{Joe_Logic} when decomposing the NLI hypothesis.} \label{Logic_Rules_Both}
\end{figure*}

We compare our training and evaluation rules to those presented by \citet{Joe_Logic} (see \cref{Logic_Rules_Both}), comparing the different approaches when either segmenting the NLI premise or hypothesis. For \citet{Joe_Logic}, when decomposing the hypothesis into atoms, entailment is the default class (when no contradiction or neutral atoms are detected), whereas when decomposing the premise into atoms in this work, neutral is the default class (if no contradiction or entailment atoms are detected).

\section{Fact Generation Strategies}
\label{sec:fact_gen_strat}

\subsection{Method}\label{sec_appendix:fact_gen_strategies_METHOD}
To maximise the likelihood that all premise information is contained in our fact list, we investigate three different fact generation strategies: 1) concatenating two independent lists of facts to reduce the likelihood that key facts are missing (we note that it should not matter if some facts are repeated), 2) we extend a given fact list to identify potentially missing facts, and 3) we generate an additional fact conditioned on the hypothesis.

When using two independent lists of facts, we generate the fact list a second time using our generator model with different examples in the prompt. These two fact lists are then combined during inference. We initially experimented with also combining both sets of fact lists during training, but we found this resulted in marginally worse performance.  In the example in \cref{fact_gen_diagram}, information about the singer is missing in the first fact list, but is included in the second fact list. Therefore, by concatenating both fact lists, we provide a more comprehensive list of facts to support our model predictions. There should be no issue with similar, duplicated facts, as the atom-level predictions should be the same for facts with identical information. 

Alternatively, we extend each fact list, using our generator model to generate additional facts to complement the initial facts already generated. In this case, the generator is presented with examples of incomplete fact lists in the prompt where the missing facts are then subsequently identified. This method aims to use our generator LLM to identify and remedy instances where key facts in the premise are missing. In \cref{fact_gen_diagram}, this method successfully includes missing information about both the album singer and the Melbourne Symphony Orchestra.

Finally, we experiment with generating premise facts conditioned on the hypothesis. This involves generating an additional fact for each example, asking the model to provide a fact explicitly known from the premise that can be used to verify if the hypothesis is true. By asking the model to produce a single fact conditioned on the hypothesis, we encourage the model to include all the relevant information in a single fact rather than requiring multi-fact reasoning. This is the case for \cref{fact_gen_diagram}, where the relevant information is condensed into a single fact that prevents the need for multi-fact reasoning.

For our fact-decomposition language model, four examples are provided in the prompt, making fact generation a few-shot task. As multiple hypotheses correspond to each premise in ANLI, not including the hypothesis in the prompt for the training data also substantially reduces the number of facts that need to be generated.

\subsection{Results from Fact Generation Strategies}\label{sec_appendix:fact_gen_strategies_RESULTS}
Out of the three fact generation strategies, we find better performance on the validation set when using hypothesis-conditioned facts (60.4\% dev accuracy), compared to using a combined fact list (58.6\% dev accuracy) and extending the fact lists (58.0\% dev accuracy). We show the test performance of each of these systems in \cref{ablation_exp_deberta_base}. The lower performance when extending a fact list can be explained by an increase in hallucinations with this strategy.

In \cref{ablation_exp_deberta_base} we additionally show performance from combining different fact-generation strategies, e.g. using combined fact lists and using hypothesis-conditioned facts, or extending the fact lists and using hypothesis-conditioned facts. These results show that combining these different strategies does not improve performance.

\section{Additional experimentation} \label{sec_appendix:additional_experimentation}
In addition to the experiments performed in the main paper using DeBERTa-base, we perform additional experiments using DeBERTa-large (see \cref{main_results_table_deberta_large}) to consider the effectiveness of our method when applied to a high-performing base model with close to state-of-the-art results. We also experiment with a BERT-base model  (see \cref{main_results_table_bert}), to consider if the findings still apply to a worse-performing base model. When applied with a DeBERTa-large model, our FGLR method significantly outperforms the FactAI baseline model for each of the ANLI in-distribution test sets (R1, R2 and R3), and also for two of the out-of-distribution test sets (ConTRoL and RTE). We also see SenLR significantly outperforming the SenAI baseline for each of the in-distribution test sets, and also for ConTRoL. When using a BERT-base baseline, FGLR is significantly better than FactAI for each ANLI dataset and also for RTE. Accuracy from SenLR is significantly better than SenAI for ConTRoL and RTE, although there is little difference in-distribution compared to the baseline.

We perform additional experiments when training on the ConTRoL dataset. These experiments include out-of-distribution evaluation on our other NLI datasets (ANLI round 1, ANLI round 2, ANLI round 3, RTE and WNLI) - see \cref{control_training_exp_deberta_base}. In this case, we find that the baseline model and FactAI do not converge (with the final model predicting only the entailment class), whereas FGLR performs better across each NLI dataset. As explained in \cref{sec:dataset_introduction}, we reduce the ConTRoL data so that no premises exceed 2,000 characters. This avoids an unfair comparison where there is truncation required for the uninterpretable base model.

Finally, we also provide the standard deviations for each result in \cref{main_results_table} in \cref{main_results_standard_deviations}.

\section{Ablation studies} \label{sec_appendix:ablation_studies}

We perform an extensive range of ablation experiments to identify the specific aspects of FGLR that are responsible for its strong performance compared to the FactAI baseline (\cref{ablation_exp_deberta_base}, \cref{ablation_exp_deberta_2}, \cref{ablation_exp_deberta_base_3a}, \cref{ablation_exp_deberta_base_4} and \cref{ablation_exp_deberta_base_5}).

First, we consider the performance of FGLR when there is either no sentence loss or fact loss (see \cref{ablation_exp_deberta_base}). These results demonstrate the importance of the fact-level loss, while the instance-level loss is responsible for a small improvement in performance. As we aim to improve the performance of the interpretable atomic inference models, we include this small performance improvement. 

We additionally experiment with using different strategies for generating the fact list for each example (see \cref{ablation_exp_deberta_base}). This includes removing the hypothesis-conditioned facts, using our strategy of combining multiple fact lists instead of using the hypothesis-conditioned facts, and combining both strategies together (using the hypothesis-conditioned facts in addition to the combined fact list). The best approach (using the hypothesis-conditioned facts) was selected based on performance on the validation set.

To understand whether the facts contain additional information that the DeBERTa-base model does not have, we also try appending these facts to the uninterpretable DeBERTa-base model. However, we find that the model does not converge in this setting (see \cref{ablation_exp_deberta_2}). This suggests that any comparison between our interpretable FGLR model and the uninterpretable DeBERTa-base model is likely to be a fair comparison.

We also consider the extent to which the FactAI baseline can be improved by including the hypothesis-conditioned facts (see \cref{ablation_exp_deberta_base_3a}). We find that improving the fact generation strategy for FactAI improves ANLI accuracy from 54.9\% to 57.4\%, almost half of the overall improvement from the complete FGLR system (60.7\% accuracy). This supports our approach of training with atoms in-the-loop, in addition to our improved fact-generation strategies.

As we utilise GPT-3 to generate our fact lists, we additionally experiment with using GPT-3 to make entailment decisions for each hypothesis and premise fact pair (see \cref{ablation_exp_deberta_base_4}). Specifically, we use a GPT-3.5-turbo model to do this in a few-shot setting (providing two examples to the model). Due to the poor performance, we concentrate our efforts on the better-performing DeBERTa models.

Finally, we provide a further experiment comparing the performance of the DeBERTa-base model to an adapted version of FGLR that does not use any atom decomposition, but instead only uses the full premise and hypothesis as inputs (See \cref{ablation_exp_deberta_base_5}). We find this system provides marginally better performance than the DeBERTa-base baseline (59.3\% compared to 58.4\%), but is still substantially lower than the performance of the full FGLR system (60.7\%).

\begin{table*}[!t]
\begin{center}
\begin{tabular}{lcccccccc}
\toprule
 
  & \multicolumn{4}{c}{\bf In-distribution} & \multicolumn{3}{c}{\bf Out-of-distribution}\\
  \cmidrule(lr){2-5} \cmidrule(lr){6-8}
  & R1 & R2 & R3 & ANLI-all & ConTRoL & RTE & WNLI & Int? \\
\midrule
DeBERTa-large & 78.3 & 66.5 & 61.7 & 68.1 & 56.0 & 90.4 & 68.9 & \xmark \\
\midrule
\textit{Span atoms:} \\
SLR-NLI & 74.7 & 60.4 & 58.3 & 64.1 & 54.7 & 
87.5 & 65.8 & \cmark \\
\midrule
\textit{Sentence atoms:} \\
SenAI & 75.3 & 63.7 & 59.1 & 65.6 & 53.4 & 86.1 & 64.7 & \cmark \\
SENTLI & 75.5 & 63.8 & 59.5 & 65.8 & 53.9 & \textbf{86.4} & \textbf{65.4} & \cmark \\
SenLR (ours) & \textbf{76.7}$\ddagger$ & \textbf{64.8}$\ddagger$ & \textbf{62.0}$\ddagger$ & \textbf{67.5}$\ddagger$ & \textbf{56.3}$\ddagger$ & 86.3 & 64.5 & \cmark \\
\midrule
\textit{Fact atoms:} \\
FactAI  & 70.0 & 60.2 & 57.3 & 62.2 & 48.3 & 81.0 & \textbf{78.7} & \cmark \\
FGLR (ours) & \textbf{76.2} $\ddagger$ & \textbf{64.8}$\ddagger$ & \textbf{63.1}$\ddagger$ & \textbf{67.7}$\ddagger$ & \textbf{52.7}$\ddagger$ & \textbf{82.0}$\dagger$ & 77.0 & \cmark \\
\bottomrule
\end{tabular}
\end{center}
\caption{Accuracy for DeBERTa-large. $\dagger$ represents results that are statistically better than the corresponding SenAI or FactAI baseline with $p < 0.05$, while $\ddagger$ represents results where $p < 0.01$, using bootstrapping statistical testing \cite{efron1993introduction}. `Int?' indicates whether the model is interpretable. All results displayed are an average from 10 different random seeds.}
\label{main_results_table_deberta_large}
\end{table*}

\begin{table*}[!t]
\begin{center}
\begin{tabular}{lcccccccc}
\toprule
 
  & \multicolumn{4}{c}{\bf In-distribution} & \multicolumn{3}{c}{\bf Out-of-distribution}\\
  \cmidrule(lr){2-5} \cmidrule(lr){6-8}
  & R1 & R2 & R3 & ANLI-all & ConTRoL & RTE & WNLI & Int? \\
\midrule
BERT-base & 54.1 & 45.7 & 45.0 & 48.1 & 47.3 & 71.2 & 49.0 & \xmark \\
\midrule
\textit{Span atoms:} \\
SLR-NLI & 51.5 & 42.5 & 42.7 & 45.4 & 46.5 & 73.9 & 44.4 & \cmark \\
\midrule
\textit{Sentence atoms:} \\
SenAI & 56.0 & 46.1 & \textbf{46.1} & \textbf{49.2} & 44.5 & 69.1 & 52.3 & \cmark \\
SENTLI & 55.4 & 46.3 & 45.9 & 49.0 & 45.5$\dagger$ & 69.8 & \textbf{53.1} & \cmark \\
SenLR (ours) & \textbf{56.1} & \textbf{46.6} & 45.7 & \textbf{49.2} & \textbf{46.9}$\dagger$ & \textbf{71.5}$\ddagger$ & 44.1  & \cmark \\
\midrule
\textit{Fact atoms:} \\
FactAI & 55.3 & 44.5 & 44.8 & 48.0 & 44.0 & 65.4 & \textbf{60.8} & \cmark \\
FGLR (ours) & \textbf{58.4}$\ddagger$ & \textbf{46.0}$\ddagger$ & \textbf{46.6}$\ddagger$ & \textbf{50.1}$\ddagger$ & \textbf{44.4} & \textbf{71.6}$\ddagger$ & 55.8 & \cmark \\
\bottomrule
\end{tabular}
\end{center}
\caption{Accuracy for BERT. \textsuperscript{1} results are reported by \cite{he2023using}. $\dagger$ represents results that are statistically better than the corresponding SenAI or FactAI baseline with $p < 0.05$, while $\ddagger$ represents results where $p < 0.01$, using bootstrapping statistical testing \cite{efron1993introduction}. `Int?' indicates whether the model is interpretable. All results displayed are an average from 10 different random seeds.}
\label{main_results_table_bert}
\end{table*}

\begin{table*}[!t]
\begin{center}
\begin{tabular}{lcccccccc}
\toprule
& \textbf{R1} & \textbf{R2} & \textbf{R3} & \textbf{ANLI-all} \\
\midrule \textit{Ablation experiments:} \\
\midrule 
FGLR - No h-cond facts & 69.6 & 54.4 & 52.4 & 58.4 \\
FGLR - No instance loss & \textbf{71.8} & 55.6 & 55.1 & 60.5 \\
FGLR - No fact loss & 42.2 & 39.2 & 37.4 & 39.5 \\
\midrule \textit{Fact generation strategies:} \\
\midrule 
FGLR - Extended fact list & 69.4 & 54.4 & 52.6 & 58.4 \\
FGLR - Combined fact list & 69.9 & 55.4 & 52.9 & 59.0 \\
FGLR - Combined fact list \& h-cond facts & 70.9 & \textbf{56.1} & 55.0 & 60.3 \\
FGLR - Extended fact list \& h-cond facts & 71.4 & 55.8 & \textbf{55.6} & 60.6 \\
\midrule 
FGLR  & \textbf{71.8} & \textbf{56.1} & 55.3 & \textbf{60.7} \\
\bottomrule
\end{tabular}
\end{center}
\caption{Ablation experiments (each an average from 10 seeds), comparing performance of our FGLR system to the following settings: 1) When the hypothesis-conditioned facts are not included (No h-cond facts) 2) the instance loss component is excluded from FGLR (No instance loss), and 3) when the fact loss component is excluded from FGLR (No fact loss). We also consider different fact generation strategies: 4) when the fact lists are extended, 5) when combining two independent fact lists, 6) when combining two independent fact lists with the hypothesis-conditioned facts, 7) when combining the extended fact list with the hypothesis-conditioned facts.}
\label{ablation_exp_deberta_base}
\end{table*}

\begin{table*}[!t]
\begin{center}
\begin{tabular}{lcccccccc}
\toprule
 & \textbf{R1} & \textbf{R2} & \textbf{R3} & \textbf{ANLI-all} \\
\midrule
DeBERTa-base baseline w/ facts appended & 33.3 & 33.3 & 33.5 & 33.4 \\
DeBERTa-base baseline & 71.2 & 54.0 & 51.7 & 58.4 \\
FGLR  & \textbf{71.8} & \textbf{56.1} & \textbf{55.3} & \textbf{60.7} \\
\bottomrule
\end{tabular}
\end{center}
\caption{We experiment with appending the generated facts to the baseline model (baseline w/ facts appended), although the model does not converge in this setting. All results are an average of 10 seeds.}
\label{ablation_exp_deberta_2}
\end{table*}

\begin{table*}[!t]
\begin{center}
\begin{tabular}{lcccccccc}
\toprule
& \textbf{R1} & \textbf{R2} & \textbf{R3} & \textbf{ANLI-all} \\
\midrule
FactAI  & 65.2 & 50.6 & 49.9 & 54.9 \\
FactAI w/ h-cond facts & 68.3 & 52.8 & 52.1 & 57.4 \\
FGLR  & \textbf{71.8} & \textbf{56.1} & \textbf{55.3} & \textbf{60.7} \\
\bottomrule
\end{tabular}
\end{center}
\caption{We experiment with providing the FactAI baseline with the hypothesis-conditioned facts, measuring the extent to which our fact-generation strategy can improve performance without training with atoms in-the-loop (FGLR). All results are an average of 10 seeds.}
\label{ablation_exp_deberta_base_3a}
\end{table*}

\begin{table*}[!t]
\begin{center}
\begin{tabular}{lcccccccc}
\toprule
& \textbf{R1} & \textbf{R2} & \textbf{R3} & \textbf{ANLI-all} \\
\midrule
FGLR w/ GPT-3.5-turbo (few-shot) & 45.0 & 39.4 & 43.8 & 42.8 \\
FGLR  & \textbf{71.8} & \textbf{56.1} & \textbf{55.3} & \textbf{60.7} \\
\bottomrule
\end{tabular}
\end{center}
\caption{We experiment with replacing our BERT or DeBERTa models in FGLR with a GPT-3.5-turbo model (in a few-shot setting, with two examples provided in the prompt). This results in poor performance compared to FGLR. All results are an average of 10 seeds.}
\label{ablation_exp_deberta_base_4}
\end{table*}

\begin{table*}[!t]
\begin{center}
\begin{tabular}{lcccccccc}
\toprule
& \textbf{R1} & \textbf{R2} & \textbf{R3} & \textbf{ANLI-all} \\
\midrule
DeBERTa-base baseline & 71.2 & 54.0 & 51.7 & 58.4 \\
FGLR w/ no facts, just full NLI prem \& hyp & \textbf{72.5} & 54.7 & 52.2 & 59.3 \\
FGLR  & 71.8 & \textbf{56.1} & \textbf{55.3} & \textbf{60.7} \\
\bottomrule
\end{tabular}
\end{center}
\caption{We train our FGLR model using only the full premise and hypothesis, without including any fact-level or sentence-level decomposition. This results in only a small improvement compared to the DeBERTa-base baseline. All results are an average of 10 seeds.}
\label{ablation_exp_deberta_base_5}
\end{table*}

\begin{table*}[!t]
\begin{center}
\begin{tabular}{lccccccc}
\toprule
 
  & \multicolumn{1}{c}{\bf In-distribution} & \multicolumn{6}{c}{\bf Out-of-distribution} \\
  & ConTRoL & R1 & R2 & R3 & ANLI-all & RTE & WNLI \\
\midrule
DeBERTa-base & 39.2 & 33.4 & 33.4 & 33.5 & 33.4 & 52.7 & 43.7 \\
\midrule
\textit{Fact atoms:} \\
FactAI & 39.2 & 33.4 & 33.4 & 33.5 & 33.4 & 52.7 & 43.7 \\
FGLR & \textbf{47.8} & \textbf{43.7} & \textbf{39.7} & \textbf{50.0} & \textbf{41.5} & \textbf{61.8} & \textbf{52.7} \\
\bottomrule
\end{tabular}
\end{center}
\caption{Training with ConTRoL, and testing on out-of-distribution NLI datasets (ANLI r1, r2, r3, RTE and WNLI). FGLR outperforms DeBERTa-base and FactAI, which both fail to converge in this setting. All results are an average of 10 seeds.}
\label{control_training_exp_deberta_base}
\end{table*}

\section{Examples of our FGLR system}\label{sec_appendix:examples}

\begin{figure*}
    \includegraphics[width=460pt]{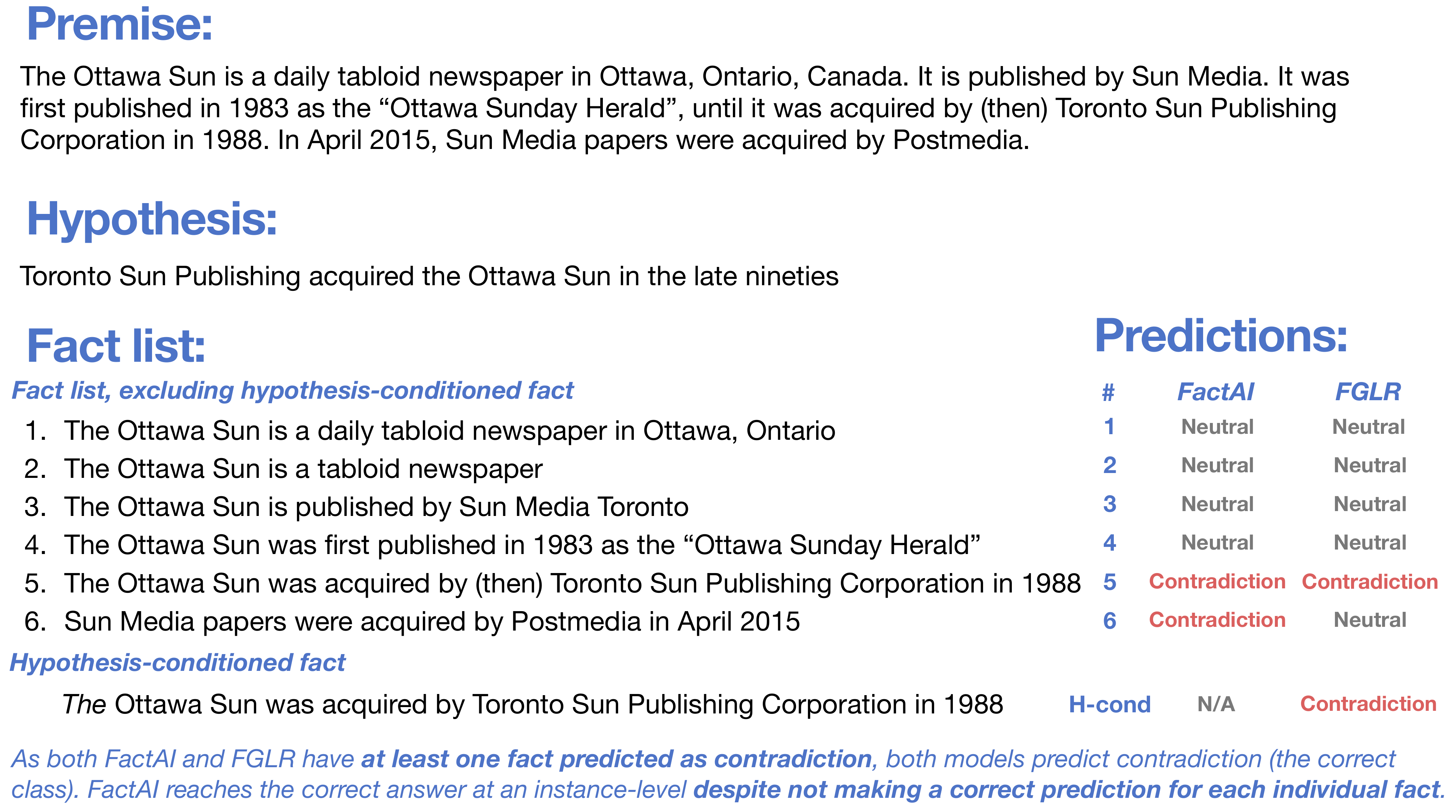} 
    \caption{We show the premise, hypothesis and the generated fact list for a hypothesis-premise pair in the dev set that both FactAI and FGLR correctly predict. However, despite correct instance-level predictions, we see FactAI predicting contradiction for fact \#6, even when this is not appropriate. In this case, the acquisition in 2015 by Postmedia does not contradict there also being an acquisition by Toronto Sun Publishing in 1988. The hypothesis conditioned fact generated for FGLR is almost identical to fact \#5, and FGLR predicts both facts as contradiction.} \label{ANLI_example_4b}
\end{figure*}

\begin{figure*}
    \includegraphics[width=460pt]{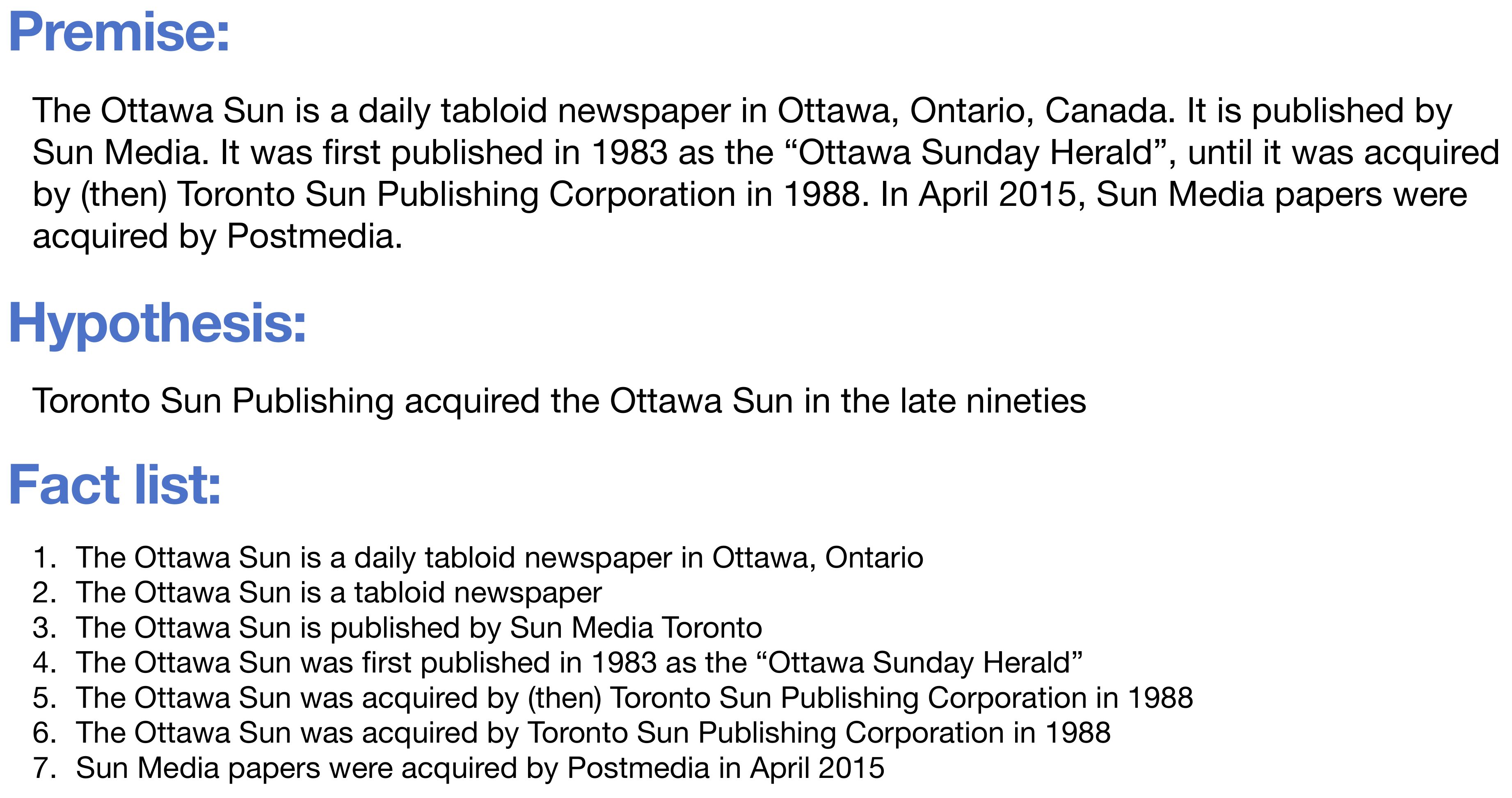} 
    \caption{The premise, hypothesis and the generated fact list for a hypothesis-premise pair in the dev set. The 6th fact is the hypothesis-conditioned fact (the content of this fact overlaps with the 5th fact provided). The example provided is from the DeBERTa-base FGLR model. FGLR correctly predicts facts 5 and 6 as contradiction (with all other facts being neutral).} \label{ANLI_example_1}
\end{figure*}

\begin{figure*}
    \includegraphics[width=460pt]{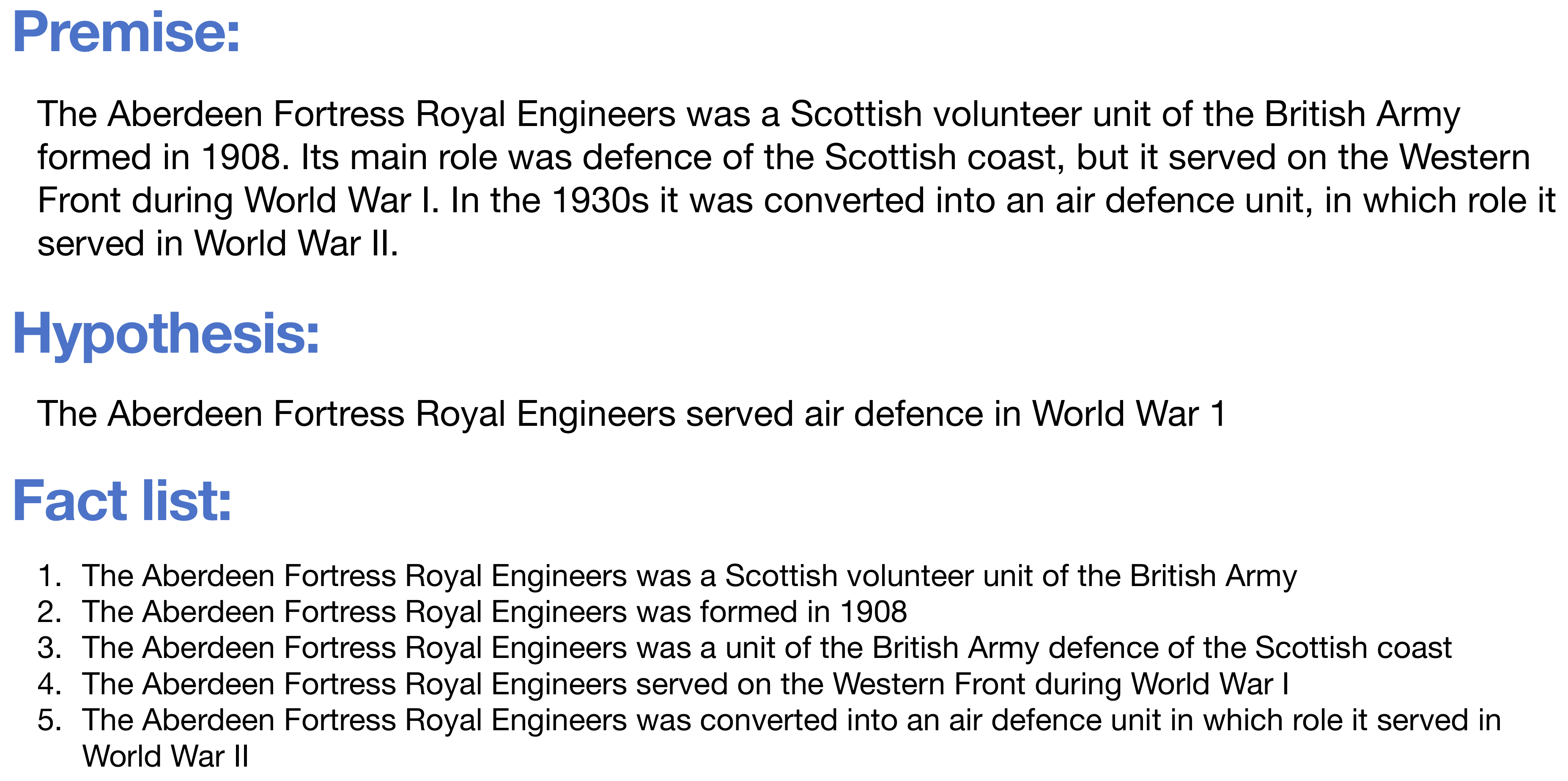} 
    \caption{The premise, hypothesis and the generated fact list for a hypothesis-premise pair in the dev set. The 4th fact is the hypothesis-conditioned fact. The example provided is from the DeBERTa-base FGLR model. FGLR correctly predicts the 5th fact as being a contradiction (with all other facts being neutral).} \label{ANLI_example_2}
\end{figure*}

We provide some examples to show the interpretability benefits of FGLR (\cref{ANLI_example_4b}, \cref{ANLI_example_1} and \cref{ANLI_example_2}). These examples are from the round 1 ANLI validation set, and consist of the generated fact-list in addition to the hypothesis-conditioned fact for each example. For \cref{ANLI_example_4b}, we compare predictions of FGLR to predictions of FactAI, showing an example where FGLR makes better predictions at a fact level.
 



\section{Hyper-parameter Tuning and Baselines}\label{sec_appendix:hyper_parameter_tuning}

For each model, we experiment with the following learning rates: $1\times10^{-6}$ to $9\times10^{-6}$ in increments of $1\times10^{-6}$, and $1\times10^{-5}$ to $9\times10^{-5}$ in increments of $1\times10^{-5}$.
The base models performed best using learning rates of $6\times10^{-5}$, $4\times10^{-5}$, and $5\times10^{-6}$, for BERT-base, DeBERTa-base, and DeBERTa-large, respectively, while the best FGLR methods used lower learning rates of $5\times10^{-6}$, $7\times10^{-6}$, and $3\times10^{-6}$, respectively. For the $\lambda$ value, we experiment with values of 0.1 to 1 in increments of 0.1, choosing a value of 0.9. Finally, we find marginally better performance if the FGLR encoder is initialised with the parameters of the fine-tuned base model. Our DeBERTa-base model consists of 184 million parameters, compared to 110 million for BERT and 304 million for DeBERTa-large. We conducted over 300 experiments, consisting of approximately 3000 GPU hours using RTX6000 GPUs.

When implementing SLR-NLI with DeBERTa-base, hypotheses with more than 50 spans were not supervised (impacting only 0.02\% training examples). This is due to memory constraints of the SLR-NLI method, which involves combining every span/premise pair into a single minibatch for each instance. This is memory intensive when hypotheses have a large number of spans and when premises are multiple sentences. For DeBERTa-large, we do not train with examples with more than 10 spans (impacting 11.46\% of instances).

\section{Detailed Comparison of FactAI and FGLR}

We find that most of the performance benefits from FGLR are from its ability to successfully distinguish between the contradiction and neutral classes. A qualitative analysis confirms that FGLR performs well in this respect, with FactAI often predicting contradiction when the information provided does not necessarily contradict the hypothesis. For example, knowing that `Judy Tegart Dalton was a runner-up in 10... tournaments' does not contradict a hypothesis that she `won more than nine... titles'. To provide empirical evidence of this finding, we try reducing the NLI task to deciding between `entailment' and `non-entailment' during inference, collapsing both the neutral and contradiction classes. In this case, we find that FactAI now outperforms FGLR on ANLI (with 73.7\% accuracy compared to 72.3\% for FGLR). This highlights how the performance advantages of FGLR are a result of its ability to successfully differentiate between the contradiction and neutral classes.

Our qualitative analysis also highlights that FGLR often predicts entailment when a fact most likely implies a premise, but when there is not full entailment. We do not see the same behaviour with the FactAI model, which is considerably less likely to predict the entailment class for an individual fact. To better understand the cause of this behaviour, we review the generated facts for 100 instances in the validation set, finding that in 21\% of cases, there is no single fact that truly implies the entire hypothesis\footnote{This increases to 46\% without the hypothesis-conditioned facts (which were not included during training)}. Sometimes this is caused by subtle reasons, for example, one hypothesis says `Shostakovich may have been lying about his life in his book', while the relevant fact says `Some consider the book Testimony to be a fabrication'. In this case, the fact is missing the information that Testimony is the name of Shostakovich's book. These findings suggest that further improvements to the generated facts are likely to further improve the performance of FGLR for entailment predictions.

We additionally analyse human annotations for each individual fact from the same 100 validation examples. To understand the respective strengths between FactAI and FGLR, we chose 2 seeds (out of 10) where both FactAI and FGLR have identical performance on these 100 examples (we also do not include the hypothesis-conditioned facts in this analysis). We find that FGLR has a lower F1 score for entailment, whereas FactAI has a lower F1 score for contradiction (see \cref{Details_for_each_class}), supporting the findings from our qualitative analysis above. 

We conclude that the performance improvements of FGLR are driven by its better performance on neutral and contradiction instances, which is reflected by better fact-level F1 performance on the contradiction class. On the other hand, we find that FactAI performs better on entailment examples, which is also reflected in the fact-level performance for the 100 annotated examples.

\begin{table*}[!t]
\begin{center}
\begin{tabular}{lcccccccc}
\toprule
 
  & \multicolumn{4}{c}{\bf In-distribution} & \multicolumn{3}{c}{\bf Out-of-distribution}\\
  \cmidrule(lr){2-5} \cmidrule(lr){6-8}
  & R1 & R2 & R3 & ANLI-all & ConTRoL & RTE & WNLI & Int? \\
\midrule
DeBERTa-base & 1.05 & 1.25 & 0.93 & 0.66 & 1.51 & 1.90 & 2.82 & \xmark \\
\midrule
\textit{Span atoms:} \\
SLR-NLI & 1.13 & 1.29 & 1.18 & 1.01 & 1.46 & 1.64 & 3.32 & \cmark \\
\midrule
\textit{Sentence atoms:} \\
SenAI & 1.29 & 1.36 & 1.10 & 0.98 & 1.53 & 1.78 & 2.30 & \cmark \\
SENTLI & 1.33 & 1.21 & 1.10 & 0.97 & 1.03 & 1.86 & 2.15 &  \cmark \\
SenLR & 0.62 & 0.96 & 0.88 & 0.60 & 1.67 & 1.34 & 2.47 & \cmark \\
\midrule
\textit{Fact atoms:} \\
FactAI & 1.47 & 1.19 & 0.81 & 0.81 & 1.74 & 1.21 & 2.82 & \cmark \\
FGLR & 0.62 & 1.08 & 0.90 & 0.54 & 1.99 & 1.10 & 2.95 & \cmark \\
\bottomrule
\end{tabular}
\end{center}
\caption{Standard deviations corresponding to the reported mean results in \cref{main_results_table} after using 10 different random seeds.}
\label{main_results_standard_deviations}
\end{table*}

\begin{table}[!t]
\begin{center}
\begin{tabular}{lcccccccc}
\toprule
& \textbf{Precision} & \textbf{Recall} & \textbf{F1} \\
\midrule
FGLR no h-cond facts \\
\midrule
Entailment & 0.24 & 0.76 & 0.36 \\
Neutral & 0.98 & 0.85 & 0.91 \\
Contradiction & 0.68 & 0.77 & 0.72 \\
Macro average & 0.63 & 0.79 & 0.66 \\ 
\midrule
FactAI \\
\midrule
Entailment & 0.44 & 0.68 & 0.53 \\
Neutral & 0.97 & 0.91 & 0.94 \\
Contradiction & 0.55 & 0.76 & 0.64 \\
Macro average & 0.65 & 0.78 & 0.70 \\
\bottomrule
\end{tabular}
\end{center}
\caption{We compare model fact-level predictions to human annotations for 100 examples in the validation set (using 2 random seeds)}
\label{Details_for_each_class}
\end{table}

\end{document}